\documentclass[12pt,a4paper,final]{amsart}

\usepackage{graphicx}
\usepackage{amssymb}
\usepackage{amsmath}
\usepackage{pifont}
\usepackage[utf8]{inputenc}
\usepackage{pifont}
\usepackage[nolist]{acronym}
\usepackage{tabularx}
\usepackage{multirow}
\usepackage{floatrow}

\newfloatcommand{capbtabbox}{table}[][\FBwidth]

\usepackage{booktabs}
\usepackage{nccmath}

\usepackage{url}
\usepackage{subcaption}
\usepackage[dvipsnames]{xcolor}
\usepackage{algorithm2e}
\usepackage{fontawesome}
\usepackage{subcaption}
\usepackage{tikz}

\usepackage{rotating}
\usepackage{pdflscape}
\usepackage{wasysym}
\usepackage{tablefootnote}

\usepackage{bookmark}

\usepackage{float}
\usepackage{hyperref}
\usepackage[noabbrev,sort&compress]{cleveref}

\begin{acronym}
    \acro{RS}{Recommender System}
    \acro{CF}{Collaborative Filtering}
    \acro{MF}{Matrix Factorization}
    \acro{MAE}{Mean Absolute Error}
    \acro{MSE}{Mean Squared Error}
    \acro{kNN}{k-Nearest Neighbors}
    \acro{CF4J}{Collaborative Filtering for Java}
    \acro{PRNG}{Pseudorandom number generator}
    \acro{PMF}{Probabilistic Matrix Factorization}
    \acro{API}{Application Programming Interface}
    \acro{NMF}{Non-negative Matrix Factorization}
    \acro{DL}{Deep Learning}
    \acro{MLN}{Multilayer Neural Network}
\end{acronym}

\author[]{Jesús Bobadilla}
\address{ETSI de Sistemas Inform\'aticos, Universidad Polit\'ecnica de Madrid. C.\ Alan Turing s/n, 28031. Madrid, Spain.}
\email{jesus.bobadilla@upm.es}
\author[]{Ra\'ul Lara-Cabrera}
\address{ETSI de Sistemas Inform\'aticos, Universidad Polit\'ecnica de Madrid. C.\ Alan Turing s/n, 28031. Madrid, Spain.}
\email{raul.lara@upm.es}
\author[]{\'Angel Gonz\'alez-Prieto}
\address{ETSI de Sistemas Inform\'aticos, Universidad Polit\'ecnica de Madrid. C.\ Alan Turing s/n, 28031. Madrid, Spain.}
\email{angel.gonzalez.prieto@upm.es}
\author[]{Fernando Ortega}
\address{ETSI de Sistemas Inform\'aticos, Universidad Polit\'ecnica de Madrid. C.\ Alan Turing s/n, 28031. Madrid, Spain.}
\email{fernando.ortega@upm.es}

\begin{document}

\newtheorem{thm}{Theorem}[section]
\newtheorem{prop}[thm]{Proposition}
\newtheorem{lem}[thm]{Lemma}
\newtheorem{cor}[thm]{Corollary}

\newtheorem{defn}[thm]{Definition}
\newtheorem{as}{Assumption}

\newtheorem{rmk}[thm]{Remark}
\newtheorem{ex}[thm]{Example}

\usetikzlibrary{bayesnet}
\usetikzlibrary{arrows}

\title[DeepFair: Deep Learning for Improving Fairness in Recommender Systems]{DeepFair: Deep Learning for Improving\\Fairness in Recommender Systems}

\date{}

    \begin{abstract}
    The lack of bias management in Recommender Systems leads to minority groups receiving unfair recommendations. Moreover, the trade-off between equity and precision makes it difficult to obtain recommendations that meet both criteria. Here we propose a \ac{DL} based \ac{CF} algorithm that provides recommendations with an optimum balance between fairness and accuracy without knowing demographic information about the users. Experimental results show that it is possible to make fair recommendations without losing a significant proportion of accuracy.\newline{}
    \null\hspace{0.5cm}\textbf{Keywords:} Recommender Systems, Collaborative Filtering, Deep Learning, Fairness, Social Equality.
    \end{abstract}

\maketitle

\section{Introduction}\label{sec:introduction}
Fairness in \ac{RS} is a very important issue, since it is part of the path to get a fair society. Nowadays, recommendations come to us from a variety of online services such as Netflix, Spotify, TripAdvisor, Facebook, Amazon, etc. All these services rely on hybrid \ac{RS}~\cite{Cano2017Jan} whose kernel is the \acf{CF}. \ac{CF} data is the set of the users' preferences on the items: tens or hundreds of millions of ratings, likes, clicks, etc. It seems great, since in theory, the more the data the better the recommendations; unfortunately, this data is usually biased~\cite{Bellogin2017Dec,Gao2020Jan} and minority groups are the most damaged ones. Common minority groups are female (vs. male) and senior (vs. young); both groups tend to receive unfair recommendations from online services. This situation has a perverse effect: a cycle that feeds back, where unfair recommendations make minority users to lose confidence in the system, to decrease their interaction and, thus, to receive even more unfair recommendations. The time has come to increase research in fair \ac{RS} as a way to reduce the digital gap~\cite{Fatehkia2018Jul,Santos2019} between minority and non-minority groups.

\ac{CF} \ac{RS} research has been traditionally focused in accuracy improvement~\cite{Portugal2018May}, although some other objectives have increased the research attention in the last years: novelty~\cite{Mendoza2020Apr}, reliability~\cite{Bobadilla2018May}, diversity~\cite{Kunaver2017May} and serendipity~\cite{deGemmis2015Sep,Kotkov2016Nov} among them. Surprisingly, fairness has not been a main objective in the \ac{RS} priorities. One of the reasons is the idea that improving fairness does not lead us to more valued recommendations, such as accuracy, novelty or diversity clearly do. Nevertheless, society needs to point in the opposite direction~\cite{Holstein2019}, and a set of new quality goals are growing~\cite{Mehrotra2018}: relevance, fairness and satisfaction among them. The historical development of \ac{CF} has not helped to the fairness research, either: when the \ac{kNN} algorithm~\cite{Herlocker2004Jan} dominated the field, it was less likely that a reduced set of neighbours produced biased recommendations. However, in a very short time the \ac{MF} method prevailed as standard, and the fairness goal relevance grew up~\cite{Hernando2016Apr}. \ac{MF} makes a compressed version of the ratings that belong to the dataset, catching the essence of them. The compressed models are sensible to the data biases such as the demographic ones: gender, age, etc.~\cite{mehrabi2019survey} making fairness a particularly relevant goal.

As a consequence of the \ac{CF} research evolution, existing publications to improve fairness using the \ac{kNN} algorithm are scarce; as an example, in~\cite{pmlr-v81-burke18a} authors look for balanced neighbourhoods as a mechanism to preserve personalization (accuracy) while enhancing the recommendations fairness. It is also remarkable the differentiation that takes place, in this context, between consumer-centred and provider-centred fairness. Fairness has been studied in the \ac{CF} context in two main directions: a) finding that data biases really generates unfair recommendations, and b) providing quality measures or methods to quantify recommendations fairness. From the first block, in~\cite{10.1145/3184558.3186949} authors argue that improving recommendations diversity leads to discrimination among the users and unfair results. The response of \ac{CF} algorithms to the demographic distribution of ratings is studied in~\cite{10.1145/3240323.3240373}; they find that common \ac{CF} algorithms differ in the gender distribution of their recommendation lists. A preliminary experimental study on synthetic data was conducted in~\cite{DBLP:journals/corr/abs-1811-01461}, where conditions under which a recommender exhibits bias disparity and the long-term effect of recommendations on data bias are investigated. From the second block (quality measures) in~\cite{DBLP:journals/corr/YaoH17} they claim that biased data can lead \ac{CF} methods to make unfair predictions for users from minority, and they propose new metrics that help reducing fairness. Disparity scores has also been proposed~\cite{10.1145/3184558.3186949} to obtain fairness measures. Bias disparity can be defined as ``how much an individual's recommendation list deviates from his or her original preferences in the training set''~\cite{DBLP:journals/corr/abs-1811-01461}, whereas average disparity measures how much preference disparity between training data and recommendation list for the minority group of users is different from that for the non-minority group~\cite{Mansoury2019Aug}. Fairness quality results in our paper implement these concepts.

Fairness in information retrieval has been focused on study data bias more than acting on the machine learning models: ``teams typically look to their training datasets, not their machine learning models, as the most important place to intervene to improve fairness in their products''~\cite{Holstein2019}. The machine learning achievements in the fairness issue have been reviewed in~\cite{DBLP:journals/corr/abs-1810-08810}, where they find some ``frontiers'' that machine learning has not crossed yet. The \ac{MF} disadvantages in \ac{CF} have been studied in~\cite{DBLP:journals/corr/YaoH17}, where authors state that the \ac{MF} model cannot manage the two main types of imbalanced data: population imbalance and observation bias. \ac{RS} fairness has been even less covered in \ac{DL} than in machine learning; as an example, in this current survey of \ac{RS} based on \ac{DL}~\cite{Mu2018Nov} the fairness goal is not mentioned, not even in its ``possible research directions'' section. The same happens with the current review paper~\cite{Batmaz2019Jun} where fairness is not mentioned despite the complete set of \ac{DL}-based \ac{RS} included in the publication. In fact, state of the art research in this area is focused on accuracy improvements~\cite{IJIMAI-3874,Bobadilla2020Jan} and it has not covered this subject. To afford a  \ac{DL}-based and fair \ac{RS} is difficult due to the neural black box model~\cite{Choo2018Jul}, that is not easy to explain or vary. Nevertheless, to tackle \ac{CF} fairness using  \ac{DL} has the advantage of providing a starting base where accuracy is high~\cite{Wu2018Apr}; it is particularly convenient since the increase in fairness usually leads to the decrease in accuracy.

For the stated reasons, the hypothesis of this paper claims that it is possible to design a  \ac{DL} architecture that provides fair \ac{CF} recommendations at the cost of reasonable decreases of accuracy. A  \ac{DL} approach to obtain fair recommendation provides a novel scenario in the \ac{RS} field. This scenario opens the door to reach accurate and fair predictions, but it is not a straightforward how to make the architectural design: we have to deal not just with raw ratings data, but also with the necessary demographic information to determine the target minority groups: female vs. male, senior vs. young, etc. Moreover, the neural network learning model cannot be changed as easily as the \ac{kNN} approach or even some machine learning algorithms. For all this, the proposed  \ac{DL} approach relies on an enriched set of input data and a tailored loss function that minimizes not only the accuracy errors but also the fairness ones. Fairness errors can be measured using the disparity scores concept~\cite{10.1145/3184558.3186949}, but how these scores are fed is a research open issue.

The proposed neural network learns from data that accomplish the current disparity concept: ``deviation from the list of recommendations and the training data''. We have specified it into two related indexes: the items one, that assigns a minority value to each item (e.g. a femininity value to a film, that depends on the female and the male preferences on this movie), and the users one, that assigns a minority value to each user (e.g. a femininity value to a user, that depends on the femininity of the items preferred for this user). Once both indexes have been set, it is possible to design a neural network loss function that rewards equality between each user minority value and his/her recommended items minority values. An additional design decision we have taken is to choose a regression approach~\cite{Bobadilla2018May} instead a classification one~\cite{Bobadilla2020Jan}: since we need to simultaneously minimize accuracy and fairness errors in the loss function, it is straightforward to pack them into a combined value so that the neural network provides us with balanced fairness/accuracy regression results. Finally, we have chosen a combined \ac{MF} and \ac{DL} approach~\cite{Bobadilla2018May,10.1145/3038912.3052569}; this design allows us to decouple the accuracy and the fairness abstraction levels by assigning accuracy to the \ac{MF} and fairness to the \ac{DL} stage.

A main advantage of the proposed architecture is that, once the model has learned, recommendations can be made to users that do not have associated demographic information; that is: we can fairly recommend to users without knowing its minority nature. It is possible because the neural network can learn the minority pattern in the same process that it learns to minimize the accuracy/fairness prediction error. It is a commercial advantage, since many users avoid filling in their personal data.  
 
The rest of the paper has been structured as follows: in~\Cref{sec:materials} the proposed method is explained and the experiments design is defined. \Cref{sec:results} shows the experiments' results and their discussions. Finally,~\Cref{sec:conclusions} contains the main conclusions of the paper and the future work.

\section{Research objective}\label{sec:research-objective}
As already discussed in \Cref{sec:introduction}, recommendation systems are primarily focused on providing recommendations with as high an accuracy as possible. This results in biased recommendations being provided to minority groups of users whose representation within the overall picture is very unbalanced. This focus on accuracy, coupled with the fact that there is a trade-off with the equity of recommendations, makes recommendations provided by a \ac{RS} focused on accuracy unfair to some minority groups.

Our research objective is to study the possibility of finding a balance between accuracy and fairness when it comes to providing recommendations to users. To this end, we propose a \ac{CF} approach capable of modulating the fairness within the recommendations.

\section{Materials and Methods}\label{sec:materials}

\begin{figure}[!h]
    \centering
    \includegraphics[width=0.8\textwidth]{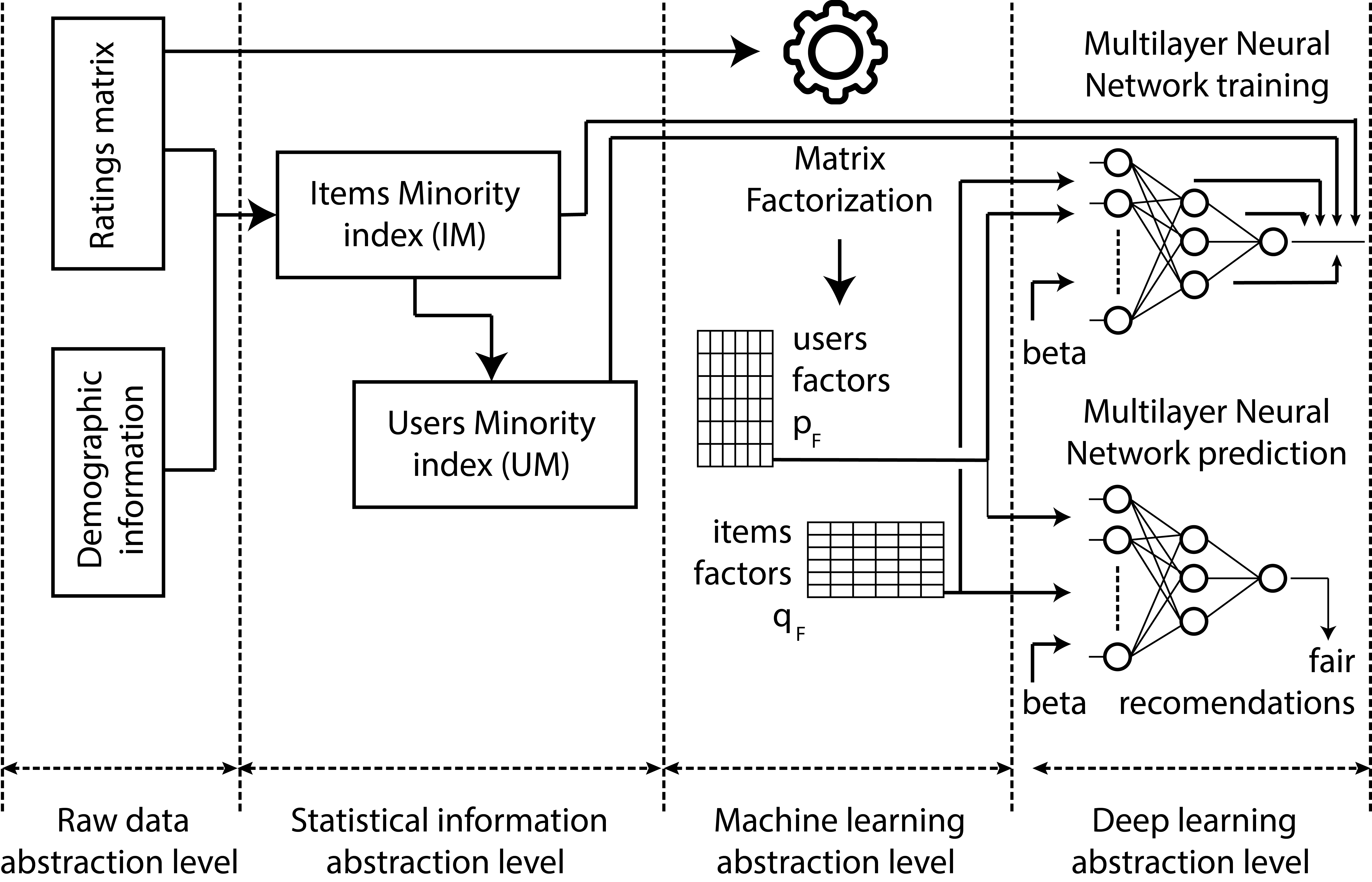}
    \caption{Architecture overview.}\label{fig:architecture}
\end{figure}

The proposed architecture incorporates four different abstraction levels, as depicted in~\Cref{fig:architecture}, to get the desired fair recommendations: a) raw ratings and demographic information, b) minority indexes for both users and items, c) accurate predictions, and d) fair recommendations. Level `b' just makes some simple statistical operations by combining ratings and demographic information; level `c' uses the classical \ac{PMF} model in order to obtain users and items hidden factors; finally, level `d' makes use of a \ac{MLN} to combine hidden factors and a `fairness' \(\beta\) parameter. This \ac{MLN} generates the desired fair recommendations.

We will develop each of the three levels that make up our architecture: first, in the lowest level we create two related indexes: 1) \emph{items minority index} (\textsc{IM}), and 2) \emph{users minority index} (\textsc{UM}). The \textsc{IM} index will assign a minority value to each item in the dataset; e.g. when the minority group is `female' we could call to the index `femininity'. It will contain values (\(\left[ -1, 1 \right]\)) where negative ones mean feminine preferences and positive ones mean masculine preferences. Then, when an item has been assigned a negative value it means that it has been rated better by women than men. Once the \textsc{IM} index has been created it contains the minority values of all the items. By using the \textsc{IM} index, we will create the \textsc{UM} index. The \textsc{UM} index will assign a minority value to each user in the dataset. It also will contain values (\(\left[ -1, 1 \right]\)), where negative ones mean minority preferences and positive ones mean not minority preferences (masculine, in our example). A user assigned a negative \textsc{UM} value means that this user prefers negative \textsc{IM} items, and vice versa. Please note that, on many occasions, female users may have assigned positive \textsc{UM} values and male users may have assigned negative \textsc{UM} values, since there exist women with masculine preferences and men with feminine ones; same as young and older persons or any other minority versus majority groups. Thus, an important concept is that both the \textsc{IM} and \textsc{UM} indexes do not contain disjoint minority/majority demographic values; they contain minority/majority preferences. This design accurately fits the existing diversity of preferences contained in the \ac{CF} based \ac{RS}.

Now, we will explain the \textsc{IM} and \textsc{UM} indexes design that we will take as a base to get fair recommendations in the \ac{DL} stage. First, we will differentiate between relevant and not relevant votes: relevant votes are those that indicate that the user liked the item; conversely not relevant votes (in our context) are those that indicate that the user did not liked the item. They can also exist votes that indicate indifference on the part of the user. In our formulation, relevant and not relevant votes are chosen by means of two thresholds; e.g. in a dataset where votes must be in the set \(\left\{ 1,2,3,4,5\right\}\) we can establish \(4\) as the relevant threshold and \(2\) as the non-relevant threshold. In this way the relevant set is \(\left\{ 5,4\right\}\), the non-relevant set is \(\left\{ 2,1\right\}\) and \(\left\{ 3\right\}\) would be the `indifference' set.

We define the \textsc{IM} index (\cref{eq:im-index}) for each item \(i\) as the majority score of \(i\) minus the minority score of \(i\). The majority score (resp. minority score) of the item \(i\) is the number of majority (resp. minority) users that voted \(i\) as relevant minus the number of majority (resp. minority) users that voted \(i\) as non-relevant, divided by the total amount of majority (resp. minority) users that did not consider \(i\) as indifferent, see \cref{eq:majority-score,eq:minority-score} (resp. \cref{eq:7,eq:8}). When the proportion of the minority user preferences exceeds the proportion of the non-minority ones, the \textsc{IM} index values are negative. In the gender example, \cref{eq:im-index} can be read as: ``proportion of males that liked item \(i\) minus males that did not like it, minus the proportion of females that liked item \(i\) minus females that did not like it''. We have also set a minimum number of \(5\) votes to consider both the minority and non-minority sides of \cref{eq:im-index}. 

Once the \textsc{IM} index has been created, we can use it to establish the UM index values. Each \textsc{UM} value corresponds to a user of the \ac{RS} dataset, and it provides the minority value of the user. Each user minority value will be defined by the minority of his/her preferences: to obtain each user \textsc{UM} value we just make the average of the \textsc{IM} minority values of the items that the user has voted, weighting each \textsc{IM} minority value with its corresponding user rating. \Cref{eq:um-index} models the explained behaviour.

\begin{flalign}
    \text{Let } &\Theta_{\uparrow} \text{ be the like threshold}\\
    \text{Let } &\Theta_{\downarrow} \text{ be the dislike threshold}\\
    \text{Let } &I \text{ be the set of items in the dataset}\\
    \text{Let } &U \text{ be the set of users in the dataset}
\end{flalign} 

We will assign the following meanings to super index numbers, \(m\) for minority and \(M\) for non-minority:

\begin{flalign}
    \text{Let } &U^m \text{ be the set of minority users}\\
    \text{Let } &U^M \text{ be the set of non-minority users}\\
    \text{Let } &U_{\uparrow}(i) = \left\{ u\in U | r_{u,i} \geq \Theta_{\uparrow}\right\} \text{ be the set of users who liked item } i\label{eq:7}\\
    \text{Let } &U_{\downarrow}(i) = \left\{ u\in U | r_{u,i} \leq \Theta_{\downarrow}\right\} \text{ be the set of users who did not like item } i\label{eq:8}
\end{flalign}

The majority score is

\begin{flalign}
    \mathfrak{I}^M(i) = \frac{|U_{\uparrow}(i)\cap U^M| - |U_{\downarrow}(i)\cap U^M|}{|U_{\uparrow}(i)\cap U^M| + |U_{\downarrow}(i)\cap U^M|}\label{eq:majority-score}
\end{flalign}

The minority score is

\begin{flalign}
    \mathfrak{I}^m(i) = \frac{|U_{\uparrow}(i)\cap U^m| - |U_{\downarrow}(i)\cap U^m|}{|U_{\uparrow}(i)\cap U^m| + |U_{\downarrow}(i)\cap U^m|}\label{eq:minority-score}
\end{flalign}

The \textsc{IM} and \textsc{UM} indexes are

\begin{flalign}
    & IM(i) = \mathfrak{I}^M - \mathfrak{I}^m\label{eq:im-index} & \\
    & IM = \left\{ \left( i, IM(i) \right)| i \in I\right\} & \\
    & UM(u) = \frac{\sum_{\left\{ i\in I | r_{u,i} \neq \circ \right\} \left( r_{u,i} - \mfrac{\Theta_1 + \Theta_2}{2}\cdot IM(i) \right)}}{\left( N - \mfrac{\Theta_1 + \Theta_2}{2} \right) \cdot | \left\{ i\in I | r{u,i} \neq \circ\right\} | }\label{eq:um-index} & \\
    & UM = \left\{ \left( u, UM(u) \right) | u\in U\right\} & 
\end{flalign}

where \(\circ\) means ``not voted item'' and \(N\) is the maximum possible vote.

\begin{figure}[!h]
    \begin{floatrow}
    \ffigbox{%
        \includegraphics[width=.4\textwidth]{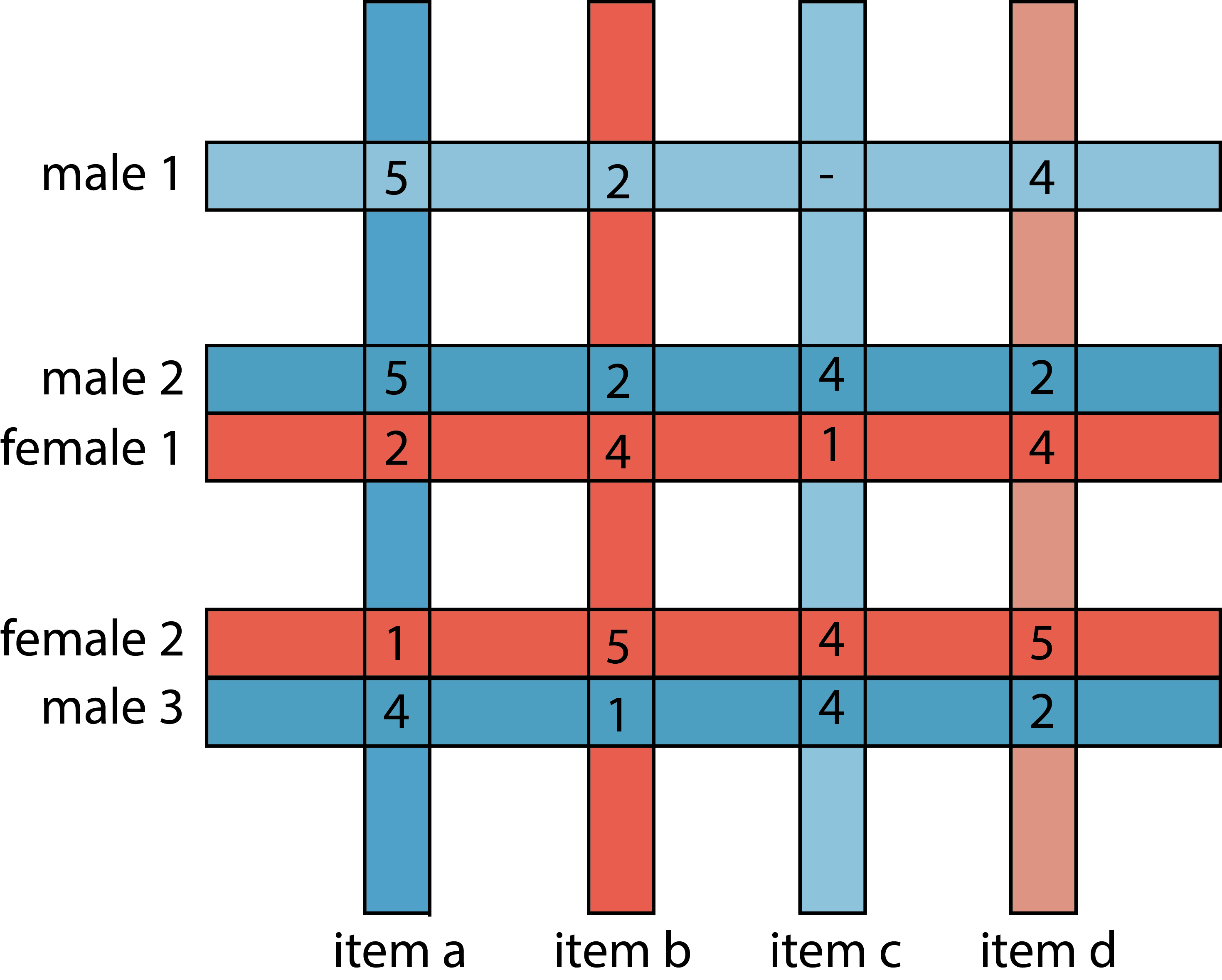}%
    }{%
    \caption{Data-toy example to get \textsc{IM} and \textsc{UM} minority values.}\label{fig:example}%
    }
    \capbtabbox{%
        \hspace{0.3cm}\resizebox{.4\textwidth}{!}{%
            \begin{tabular}{lr}
                \toprule
                item & value                  \\ \midrule
                a    & $[(3-0)-(0-2)]/5 = 1$\\
                b    & $[(0-3)-(2-0)]/5 = -1$\\
                c    & $[(2-0)-(1-1)]/4 = 0.5$\\
                d    & $[(1-2)-(2-0)]/5 = -0.6$ \\ \bottomrule
            \end{tabular}%
        }%
    }{%
      \caption{Data-toy \textsc{IM} results}\label{tbl:example}%
    }
    \end{floatrow}
\end{figure}

\Cref{fig:example} shows a data-toy example containing five users and four items. We will suppose that women are a minority group in this \ac{RS}, compared to the men. We can observe that `\emph{item a}' is clearly `masculine', since it has been voted as `relevant' for all the male users and it has been voted as `non-relevant' for all the female users. The opposite situation is stated in `\emph{item b}': it is a `feminine' item according to the female relevant votes and the male non-relevant ones. `\emph{Item c}' is quite masculine, although a female user liked it. Finally, `\emph{item d}' shows the opposite situation to `\emph{item c}'. According to it, the proposed \textsc{IM} equations return the following item minority values: 
$$
    \left\{ \left \langle \text{item a},1 \right \rangle, \left \langle \text{item b},-1 \right \rangle, \left \langle \text{item c},0.5 \right \rangle, \left \langle \text{item d},-0.6 \right \rangle \right\}
$$
that fits with the explained behaviour (\Cref{tbl:example}). Once the items' minority values \textsc{IM} are obtained, we can get the users minority ones (\textsc{UM}). First, we can observe how `\emph{male 2}' and `\emph{male 3}' users in the data-toy example have casted very `masculine' ratings, since they have voted `relevant' to the more `masculine' items, and `non-relevant' to the more `feminine' items. This is not the case for the `\emph{male 1}' user, that has a `relevant' vote casted on the `feminine' `\emph{item d}'. The female users comparative is more complicated: `\emph{female 1}' has casted all her votes in a `feminine' way, whereas the `\emph{female 2}' vote to the `masculine' `\emph{item c}' was `relevant'; nevertheless, the `\emph{female 2}' feminine votes are higher than the `\emph{feminine 1}' ones. In this way, we expect the following results: a) positive \textsc{UM} values to male users and negative ones to female users, and b) a more `minority' (feminine) value be assigned to `\emph{male 1}' than to `\emph{male 2}' and `\emph{male 3}'. \Cref{tbl:example} shows the \Cref{fig:example} data-toy \textsc{IM} results and \Cref{tbl:um-example} shows the \textsc{UM} ones.

\begin{table}[ht]
    \centering
    \resizebox{\textwidth}{!}{%
    \begin{tabular}{@{}lr@{}}
    \toprule
    user     & value                                                         \\ \midrule
    male 1   & \((5-3)\cdot1 + (2-3)\cdot(-1) +   (4-3)\cdot(-0.6) = 2.4 /5 = 0.48         \)\\
    male 2   & \((5-3)\cdot1+(2-3)\cdot(-1)+(4-3)\cdot0.5+(2-3)\cdot(-0.6)   = 4.1 / 5 = 0.82  \)\\
    female 1 & \((2-3)\cdot1+(4-3)\cdot(-1)+(1-3)\cdot0.5+(4-3)\cdot(-0.6)   = -3.6/ 5 = -0.72 \)\\
    female 2 & \((1-3)\cdot1+(5-3)\cdot(-1)+(4-3)\cdot0.5+(5-3)\cdot(-0.6)   = -4.7/ 5 = -0.94 \)\\
    male 3   & \((4-3)\cdot1+   (1-3)\cdot(-1)+(4-3)\cdot0.5+(2-3)\cdot(-0.6) = 4.1/ 5 = 0.82  \)\\ \bottomrule
    \end{tabular}%
    }
    \caption{Data-toy UM results}
    \label{tbl:um-example}
\end{table}

Our architecture uses the \ac{PMF} method to reduce the ratings matrix dimension and to get a condensed knowledge representation. From the condensed results we will be able to make accurate predictions. \Crefrange{eq:15}{eq:25} show the model formalization: the original ratings matrix is condensed in the two lower dimension matrices \(P\) and \(Q\) (\cref{eq:15}). \(P\) is the users' matrix and \(Q\) is the items' matrix. Both \(P\) and \(Q\) have a common dimension of \(F\) hidden factors, where \(F \ll M\) and \(F \ll M\) (note that \(M\) is numbers of users, and \(N\) the number of items). Once the model has learnt, each user will be represented by a vector \(\vec{p}_u\) of \(F\) factors, and each item will be also represented by a vector \(\vec{q}_i\) of \(F\) factors. Each prediction of an item \(u\) to a user \(i\) is obtained by processing the dot product of these vectors (\cref{eq:17}). Since the users and the items hidden factors share the same semantic, predictions will be relevant when high values (positive or negative) of the factors line up in each user and item.

\begingroup
\allowdisplaybreaks
\begin{flalign}
    & R \approx \hat{R} = P \cdot Q^t\label{eq:15} & \\
    & \hat{r}_{u,i} = \vec{p}_u \cdot \vec{q}_i\label{eq:17} = \sum_{f=1}^F p_{u,f} \cdot q_{i,f} & 
\end{flalign}
\endgroup
The \(P\) and \(Q\) factors will be used in our architecture to feed the \ac{DL} process input as well as to set the output target labels. Factors are obtained by means of the gradient descent algorithm. The loss function just minimizes the prediction error: the difference between the predicted value and the existing rating (\cref{eq:18}).

\begingroup
\allowdisplaybreaks
\begin{flalign}
    & \mathrm{loss}(u,i) = \left( r_{u,i} - \hat{r}_{u,i} \right)^2 \label{eq:18} & 
\end{flalign}
\endgroup

In order to achieve the gradient descent minimization process we obtain the partial loss derivatives: \(\delta \mathrm{loss} / \delta \vec{p}_u\) and \(\delta \mathrm{loss} / \delta \vec{q}_i\) (\cref{eq:19,eq:20}). 

\begingroup
\allowdisplaybreaks
\begin{flalign}
    & \frac{\delta \mathrm{loss}}{\delta \vec{p}_u} = \frac{\delta}{\delta \vec{p}_u} \left( r_{u,i} - \vec{p}_u \cdot \vec{q}_i \right)^2 = -2\vec{q}_i \cdot \left( r_{u,i} - \vec{p}_u \cdot \vec{q}_i \right) = -2\vec{q}_i \cdot e_{u,i}\label{eq:19} & \\
    & \frac{\delta \mathrm{loss}}{\delta \vec{q}_i} = \frac{\delta}{\delta \vec{q}_i} \left( r_{u,i} - \vec{p}_u \cdot \vec{q}_i \right)^2 = -2\vec{p}_u \cdot \left( r_{u,i} - \vec{p}_u \cdot \vec{q}_i \right) = -2\vec{p}_u \cdot e_{u,i}\label{eq:20} & 
\end{flalign}
\endgroup

This gives rise to the corresponding gradient descent factors update \Cref{eq:21,eq:22}.
\begingroup
\allowdisplaybreaks
\begin{flalign}
    & \acute{p}_{u,f} = p_{u,f} + 2\gamma \cdot q_{f,i} \cdot e_{u,i}\label{eq:21} & \\
    & \acute{q}_{f,i} = q_{f,i} + 2\gamma \cdot p_{q,f} \cdot e_{u,i}\label{eq:22} & 
\end{flalign}
\endgroup

Finally, we can add a regularization term for controlling the growing of the factors during the learning process, which gives rise to the loss function and the update rules shown in \Crefrange{eq:23}{eq:25}.

\begingroup
\allowdisplaybreaks
\begin{flalign}
    & \mathrm{loss}(u,i) = \left( r_{u,i} - \hat{r}_{u,i} \right)^2 + \frac{\lambda}{2} \sum_{f=1}^{F}\left( \left \| P^2 \right \| + \left \| Q^2 \right \| \right)\label{eq:23} & \\
    & \acute{p}_{u,f} = p_{u,f} + \gamma \left( 2q_{f,i} \cdot e_{u,i} - \lambda \cdot p_{uf} \right)\label{eq:24} & \\
    & \acute{q}_{f,i} = q_{f,i} + \gamma \left( 2p_{uf} \cdot e_{u,i} - \lambda \cdot q_{f,i} \right)\label{eq:25} &
\end{flalign}
\endgroup

\begin{figure}[ht]
    \centering
    \includegraphics[width=\textwidth]{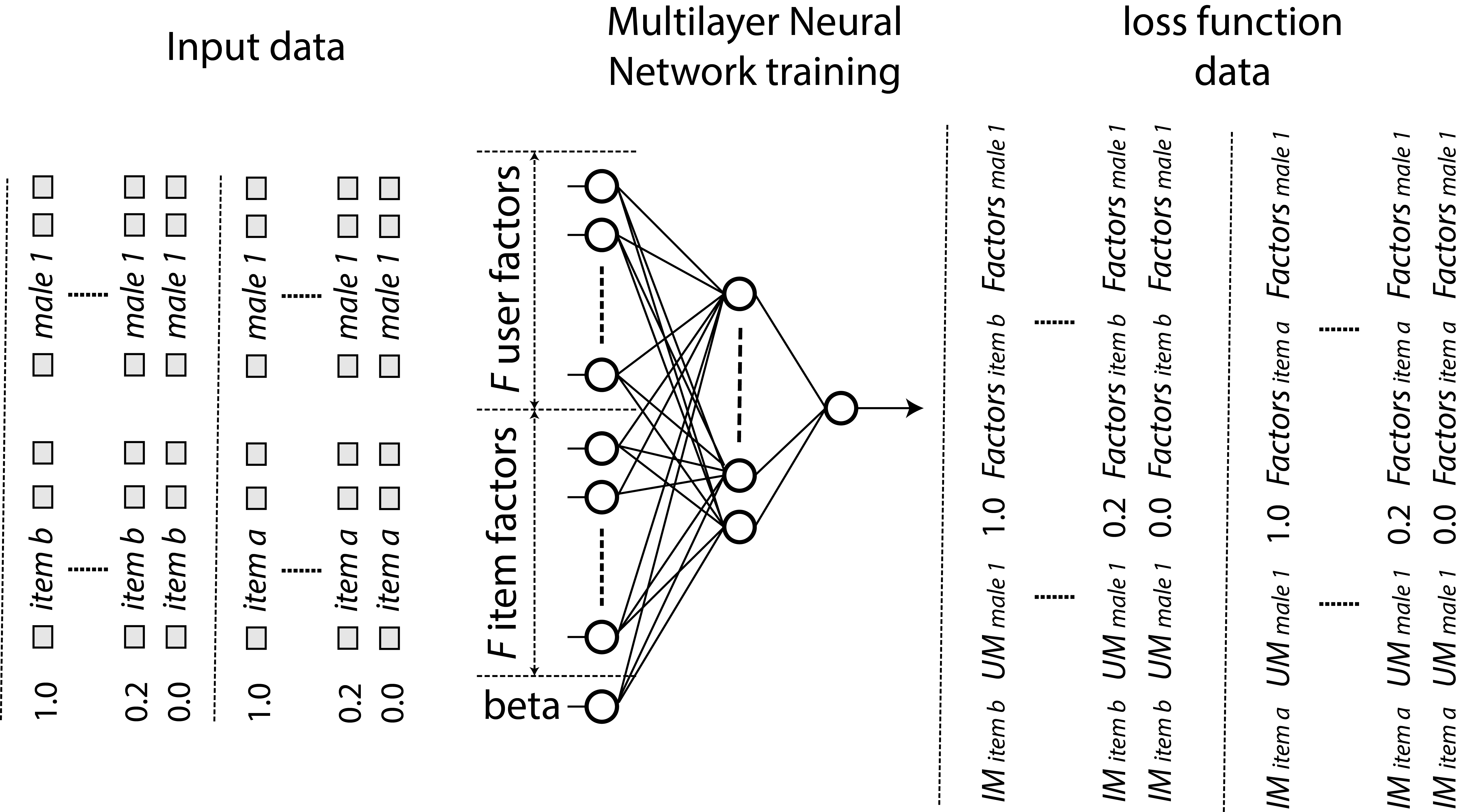}
    \caption{Training information for the proposed MLN.}\label{fig:training}
\end{figure}

The highest semantic level of the proposed architecture is based on an \ac{MLN}. Our \ac{MLN} (see \cref{fig:training}) model will take input vectors containing the following information: a) user hidden factors \(\vec{p}_u\), b) item hidden factors \(\vec{q}_i\), and c) \(\beta \in \left[ 0,1 \right]\) value. The \(\beta\) parameter is used to balance fairness and accuracy in predictions and recommendations: high \(\beta\) values will enhance accuracy, whereas low \(\beta\) values will enhance fairness. This balance is a key objective of our method: ``To obtain fair recommendations just losing an acceptable degree of accuracy''. Please note that we do not include demographic information to feed the \ac{MLN} input, so once the \ac{MLN} has learnt it will be able to make fair recommendations to users that have not filled demographic forms asking for gender, age, etc. This is an important commercial advantage, since it allows to make better marketing processes, to improve fairness, to focus prediction tasks, etc. It is also a challenge to the proposed machine learning framework, because it is more difficult to increase recommendation fairness when demographic data is missing. The learning process has been based on input vectors containing the specified three information sources:\(\left\langle p_{u,f},q_{f,i},\beta \right\rangle\) . We have set 11 input vectors to the \ac{MLN} for each \(\left\langle \text{user u}, \text{item i}\right\rangle\) rating of the dataset:

$$
\left\langle p_{u,f},q_{f,i},0.0\right\rangle, \left\langle p_{u,f},q_{f,i},0.2\right\rangle, \ldots, \left\langle p_{u,f},q_{f,i},1.0\right\rangle
$$

The objective is to teach to the neural network on eleven fairness levels for each rating, as it can be seen in the left side of \cref{fig:training}.

Once the \ac{MLN} input vectors have been established, it is necessary to define their corresponding output labels in order to let the back propagation algorithm learn the pattern. In our case we will design a loss function that minimizes both the prediction error and the fairness error. \Cref{eq:26} shows the typical prediction loss function, as we did in \cref{eq:18}. We define the fairness error as the distance between the user's minority and the item's minority; e.g. films recommended to a user (male or female) with an assigned \(-0.8\) \textsc{UM} femininity value should be as similar as possible to a \(-0.8\) \textsc{IM} in order to fit in the fairness issue. Since \textsc{UM} and \textsc{IM} vector values do not have the same distribution, we will apply a \(\left [0,1 \right]\) normalization in both of them and we will use the \textsc{UM}’ and \textsc{IM}’ names for the normalized versions. Then, to obtain the fairness error we establish \cref{eq:27}. Finally, to combine \cref{eq:26} (accuracy) and \cref{eq:27} (fairness) the \(\beta\) parameter is added (\cref{eq:28}).

\begin{flalign}
    & e_{u,i}^{\mathrm{accuracy}} = \left( r_{u,i} - \sum_{f=0}^{F} p_{u,f} \cdot q_{i,f} \right)^2\label{eq:26} & \\
    & e_{u,i}^{\mathrm{fairness}} = \left( IM'_i - UM'_u \right)^2\label{eq:27} & \\
    & \mathrm{loss}_{u,i} = \beta \cdot e_{u,i}^{\mathrm{accuracy}} + \left( 1 - \beta \right)\cdot e_{u,i}^{\mathrm{fairness}}\label{eq:28} &
\end{flalign}

In the feed forward prediction stage, for each testing input data \(\left\langle p_{u,f},q_{f,i}, \beta \right\rangle\), the proposed neural network returns a real number whose meaning is the predicted loss error for the item \(i\) to the user \(u\) recommendation. The lower the predicted loss error, the better the combined \(\left\langle \mathrm{accuracy}, \mathrm{fairness}\right\rangle\) values given the chosen \(\beta\) accuracy vs. fairness balance. Once the network has learnt and the \ac{RS} is in production phase, to make recommendations to an active user \(u\), first we fix the \(\beta\) value and then we feed the \ac{MLN} with all the inputs \(\left\langle \vec{p}_u, \vec{q}_i, \beta \right\rangle\) where \(i\) runs over the set of items that the user \(u\) has not voted (\cref{eq:29}).

\begin{flalign}
    & X = \left\{ \left\langle p_{u,f},q_{f,i},\beta \right\rangle \Big| u\in U, i\in I, r_{u,i}\neq \circ \right\}\label{eq:29} &
\end{flalign}

The set of \(N\) recommendations for the user \(u\), \(Z_{u,N}\) is the collection of \(N\) items with minimum loss function \(h(\vec{p}_u,\vec{q}_i,\beta)\), where the \(h\) function represents the \ac{MLN} feed forward operation.

Experiments have been conducted using a well known dataset called \textsc{MovieLens 1M}~\cite{10.1145/2827872}. It contains 1,209,000 votes, 6040 users and 3952 items. We have used eleven different values of the \(\beta\) parameter (from 0.0 to 1.0, step 0.2); consequently, the \ac{MLN} has been trained using 13,299,000 input vectors and output target values. Training, validation and test sets have been established: 70\%, 10\% and 20\%, respectively. The \ac{PMF} process has been run using 30 hidden factors (\(F\)), 80\% training ratings, 20\% testing ratings. Please note that these are the \ac{MLN} parameters of the proposed method, different to the previously ones specified for the \ac{DL} stage. The designed \ac{MLN} contains an input layer of \(30+30+1=61\) values (\cref{fig:training}). The first \ac{MLN} internal layer has been set to 80 neurons (\texttt{relu} activation), followed by a 0.2 dropout layer to avoid overfitting. The second internal layer has been set to 10 neurons (\texttt{relu} activation) and, finally, the output layer contains just one neuron with no activation function. The chosen loss function has been \texttt{mae} and the optimizer \texttt{rmsprop}.

\section{Results}\label{sec:results}

The experiments we have conducted are:
\begin{itemize}
    \item Item Minority Index (\textsc{IM}) and User Minority Index (\textsc{UM}) distributions.
    \item User Minority Index (\textsc{UM}) comparative between each minority and non-minority group.
    \item Fairness prediction improvement using the heuristic algorithm.
    \item Fairness recommendation improvement using the heuristic algorithm.
    \item Fairness error and accuracy error for recommendations using the proposed \ac{DL} architecture.
\end{itemize}

\begin{figure}[ht]
    \centering
    \includegraphics[width=.5\textwidth]{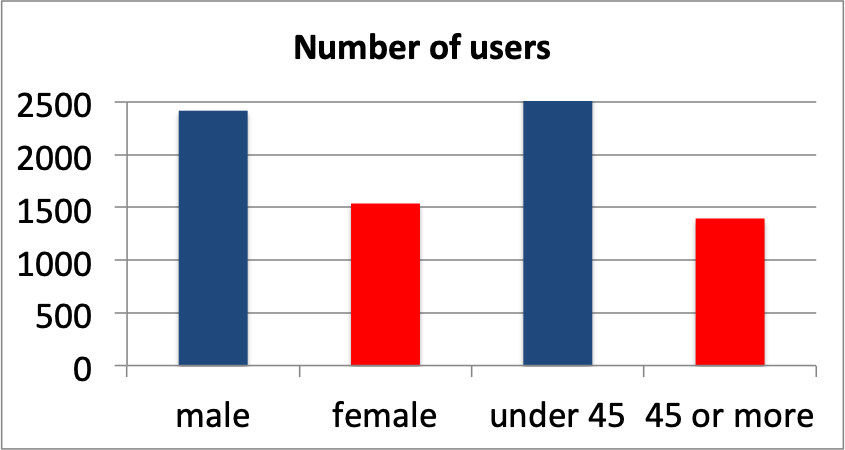}
    \caption{Proportion of users in the MovieLens gender and age minority and non-minority groups.}\label{fig:proportion}
\end{figure}

This section contains a subsection for each of the above set of performed experiments. We have selected two types of minority sets: a) gender: female vs. male, and b) youth: young vs. senior. Results are provided showing both minority types in two separated graphs of each figure. The MovieLens dataset, like in many other \ac{CF} \ac{RS} happens, is biased towards female and young people. Thus, the chosen minority types are relevant and representative for this experimental study. Specifically, the \textsc{MovieLens} dataset contains more males than females; most of them are under 45 years old. \Cref{fig:proportion} shows the proportions.

\begin{figure}[ht]
    \centering
    \includegraphics[width=\textwidth]{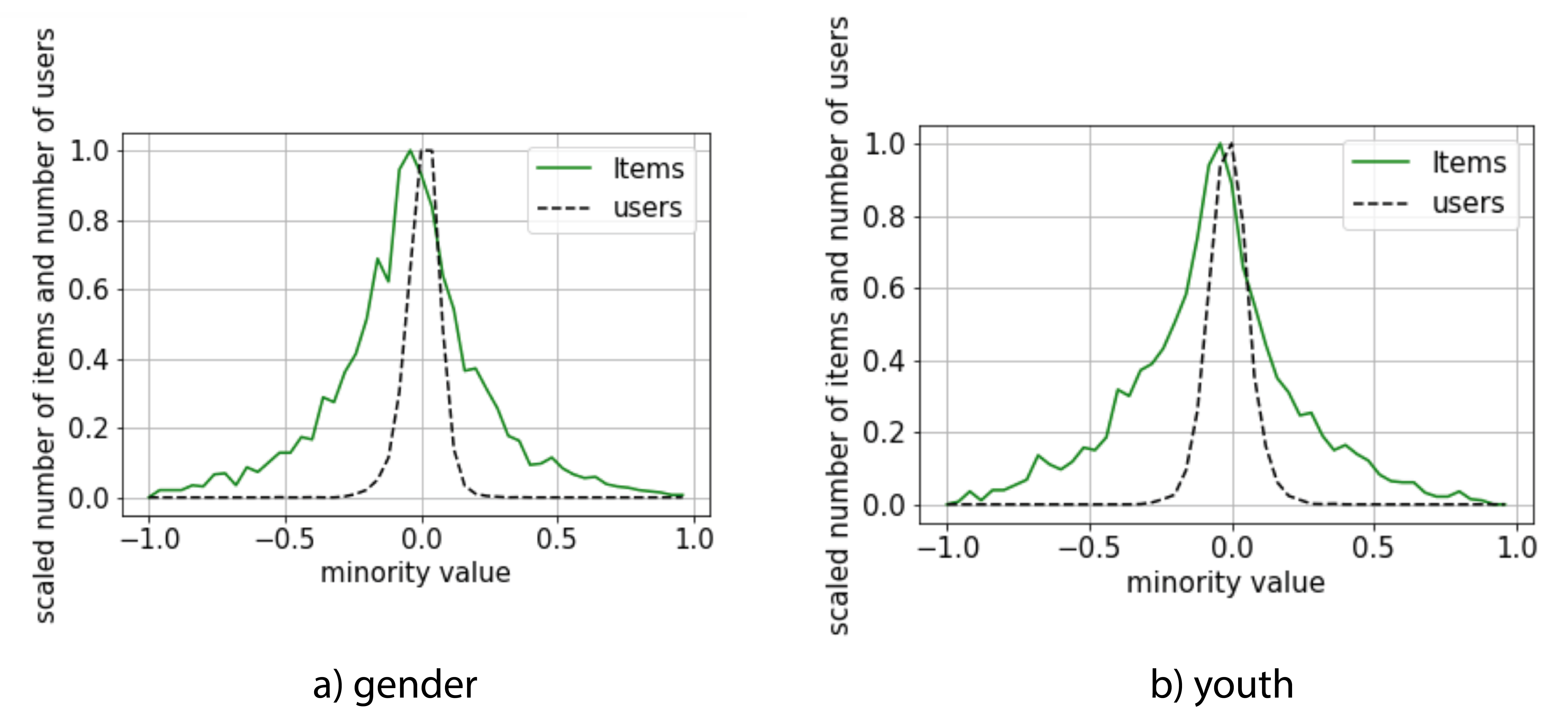}
    \caption{Item Minority Index (\textsc{IM}) and User Minority Index (\textsc{UM}) distributions.}\label{fig:distributions}
\end{figure}

\Cref{eq:im-index,eq:um-index} describe both indexes behaviour. The \textsc{IM} index semantic is simple and convincing, but it is necessary to be aware that we are not working with absolute values: in order to prevent data biases and to maintain the index values in a bounded range, we are working with preferences proportions; e.g. ``proportion of male users that liked the items minus proportion of female users that liked the item''. Since we expect a significant number of items that both minority and non-minority groups simultaneously like or dislike, \textsc{IM} proportions will be similar for both groups and consequently a significant number of \textsc{IM} values will concentrate around the 0.0 value. \Cref{fig:distributions} shows the items and users minority indexes distributions, both for the gender and the youth minority groups.

The \textsc{UM} index values are obtained from the ratings that each user has casted to the items and from the \textsc{IM} value of each of those items. We can see in \cref{fig:distributions} that the users \textsc{UM} indexes (both for gender and youth) have a large concentration of values around 0. It provides us an important conclusion: ``In the reference dataset, most users have similar preferences regarding to the chosen minority groups''. Looking at the \textsc{UM} distributions we can also yield another main conclusion: ``Although users have similar preferences, there is a clear separation between minority groups'' (left and right side of the graphs). Since the \textsc{UM} index is only used to feed internal \ac{DL} processes the relevant information here is the proportion of the differences between values, and not their absolute values.

\subsection{User Minority Index (UM) comparative between each minority and non-minority group}

\begin{figure}[ht]
    \centering
    \includegraphics[width=\textwidth]{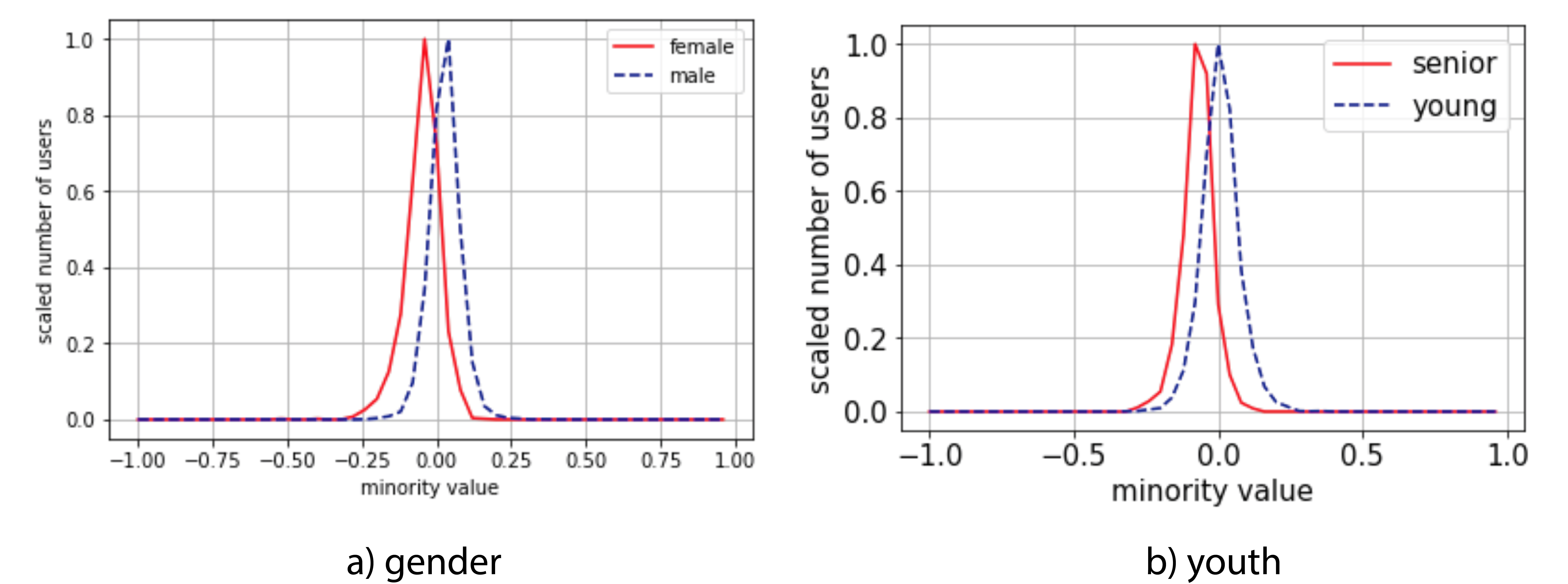}
    \caption{User Minority Index (UM) comparative.}\label{fig:um-comparative}
\end{figure}

In the above section we have confirmed two facts: 1) Users preferences are similar, even if they belong to different minority groups, and 2) Despite the previous conclusion, there is room to find minority behaviours of users. In this section we deepen in the minority \textsc{UM} values of users, to clear out our specific groups: male vs. female and senior vs. young. \Cref{fig:um-comparative} shows the results: we can observe, in both cases, that groups have different behaviours and also that they share a relevant number of preferences. Groups present different behaviours because they do not completely intersect their user minority values; as expected, minority groups return a mean less than zero whereas non-minority groups return it greater than zero. Groups share a relevant number of preferences because there exist a proportion of minority and non-minority users that share \textsc{UM} values (areas around 0.0 under both curves).

\begin{table}[ht]
    \centering
    \resizebox{.6\textwidth}{!}{%
    \begin{tabular}{@{}lllll@{}}
    \toprule
    group                   & type   & correct & incorrect & correct\% \\ \midrule
    \multirow{2}{*}{gender} & female & 1147    & 562       & 67.11     \\
                            & male   & 3648    & 683       & 84.22     \\
    \multirow{2}{*}{youth}  & senior & 1231    & 195       & 86.32     \\
                            & young  & 3144    & 1470      & 68.14     \\ \midrule
    \end{tabular}%
    }
    \caption{Users classification attending to the minority/non-minority groups.}
    \label{tab:classification}
    \end{table}

Due to the explained results, we can confirm that there is a not negligible proportion of minority users with non-minority preferences and vice versa. In any case, it varies depending on the specific minority group. As an example, we can observe in \cref{fig:um-comparative} how senior users have much less non-minority preferences than female ones, since there are small amounts of senior users whose minority value is greater than zero. Results show the convenience of using modern machine learning approaches to make fair recommendations to those users that share minority and non-minority preferences. \Cref{tab:classification} shows the specific number of users that have been classified as belonging to the minority or to the non-minority groups. Minority users (female, young) have an expected UM index less than zero. Non-minority users (male, senior) have an expected \textsc{UM} index greater than zero.

\subsection{Fairness prediction improvement using a heuristic algorithm}

\begin{table}[ht]
    \centering
    \resizebox{0.5\textwidth}{!}{%
    \begin{tabular}{@{}lllll@{}}
    \toprule
            & female & male  & senior & young \\ \midrule
    IM mean & -0.014 & 0.041 & -0.025 & 0.028 \\ \bottomrule
    \end{tabular}%
    }
    \caption{Averaged \textsc{IM} values for the predictions made to each users’ group}
    \label{tab:im-predictions}
\end{table}

\Cref{fig:um-comparative} and \cref{tab:im-predictions} show us that the majority of the users are correctly grouped attending to their \textsc{UM} indexes, especially for seniors and males. They also show a considerable number of cases incorrectly classified, particularly for young and female groups. In this situation, we will obtain predictions from the test set and then check their quality in terms of the \textsc{IM} index. \Cref{tab:im-predictions} contains these experiments results: the \textsc{IM} averages fit the expected ranges (negative \textsc{IM} average for minority users, and positive \textsc{IM} average for non-minority users). Despite these positive results, ranges can be too narrow to ensure fair predictions. On the other hand, there will be situations in which it is intended to force the recommendations of an \ac{RS} to move towards minority items, or perhaps towards majority items, depending on the type of users and/or the company policy.

\begin{figure}[ht]
    \centering
    \includegraphics[width=\textwidth]{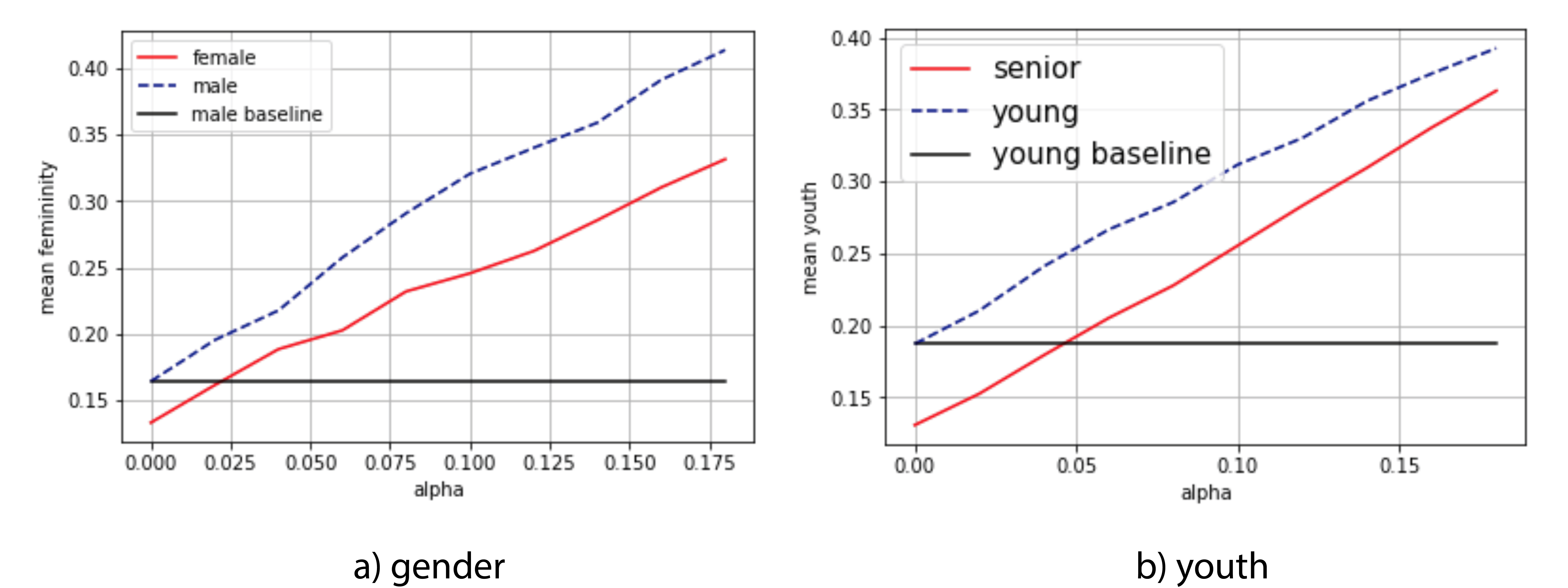}
    \caption{Groups quality improvement by filtering predictions. x axis: alpha values used to filter on the \textsc{IM} items index. y axis: averaged minority of the filtered predictions. Minority (female, senior) curves are drawn using their absolute values.}\label{fig:groups}
\end{figure}

By filtering on the \textsc{IM} index, we can discard those predictions greater than a negative threshold and, in this way, increase the proportion of minority predictions. In the same way we can filter those predictions less than a positive threshold to increase the proportion of majority predictions. We have performed this experiment, calling alpha to the threshold. We can observe the expected behaviour in \cref{fig:groups}, where growing minority (and majority) \textsc{IM} values are obtained in predictions when the alpha parameter increases. It also can be seen that the non-minority users (male, young) always obtain better predictions due to the \ac{RS} datasets biases. Finally, we can state that, in this case, minority values can reach the starting majority ones by using low values of the alpha parameter (0.025 for gender and 0.05 for age).

\subsection{Fairness recommendation improvement using the heuristic algorithm}

The previous section results show that it is possible to provide a heuristic method to improve recommendations fairness. To conduct the experiment, from the alpha filtered predictions (\Cref{fig:groups}), we extract the \(N\) ones that provide higher prediction values, as usual in the \ac{CF} operation.  Thus, the complete recommendation method involves three sequential phases: 1) to obtain all the \(\left\langle \text{prediction value}, \text{minority value}\right\rangle\) pairs, 2) to filter the pairs according to the minority threshold alpha parameter and each \emph{minority value}, and 3) to select the \(N\) filtered predictions that have the \(N\) highest \emph{prediction value} values.

\begin{figure}[ht]
    \centering
    \includegraphics[width=\textwidth]{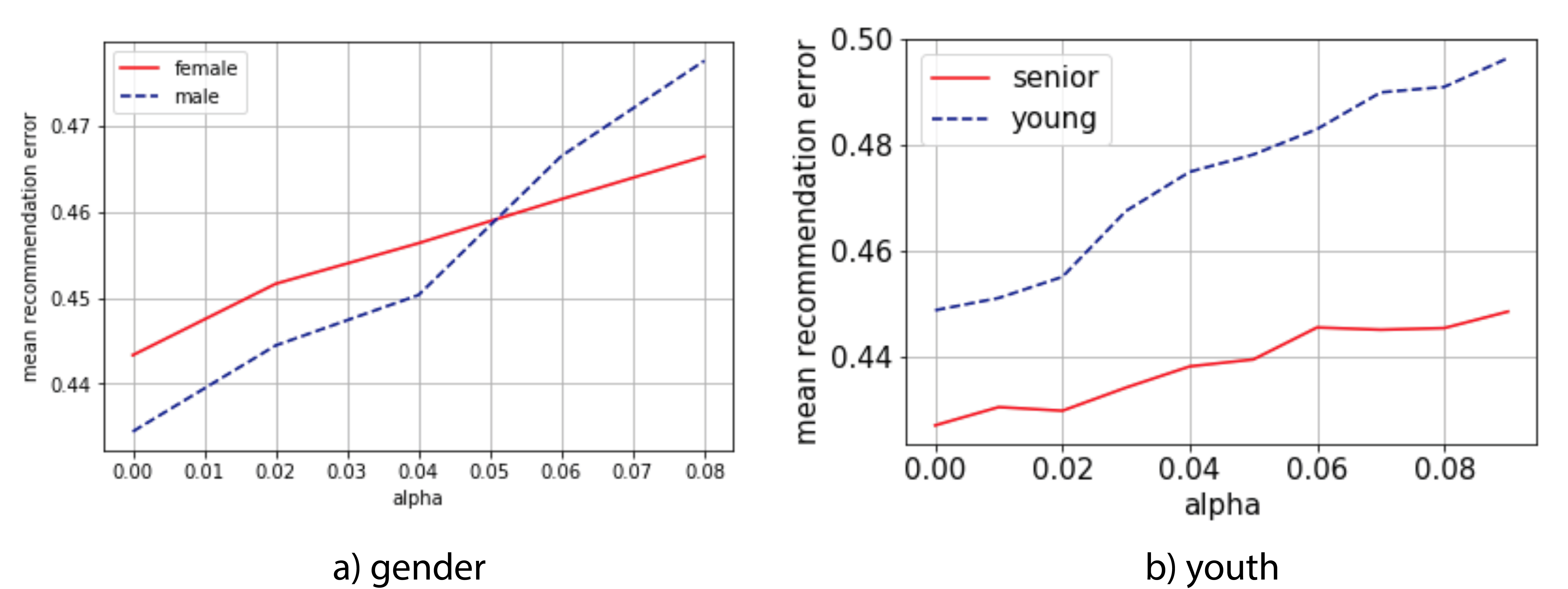}
    \caption{Recommendation quality obtained by filtering predictions. x axis: alpha values used to filter on the \textsc{IM} items index. y axis: averaged error of the \(N\) recommendations. Lower error values are the better ones.}\label{fig:rec-quality}
\end{figure}

Results in \cref{fig:rec-quality} show the existing correlation between recommendation errors and each chosen alpha value: the highest the alpha value, the better the recommendations fairness (\Cref{fig:groups}), but as expected, also the worst the recommendation accuracy (higher error values in \cref{fig:rec-quality}). Of course, we pay an accuracy price when we force fairer recommendations.

We have chosen a value of \(N=10\) recommendations to process the set of experiments. From \cref{fig:groups} it can be observed that in the ‘youth’ experiment our method provides better results (lower errors) for the minority ‘senior’ group than for the ‘young’ one. This is a good indication of the proposed heuristic method functioning. The ‘gender’ experiments provide improvement in the minority female group from a specific value threshold (\(\mathrm{alpha} = 0.05\)). All these results are consistent with \Cref{tbl:um-example,tab:classification} values.

\subsection{Fairness error and accuracy error for recommendations using the proposed \ac{DL} architecture}

Results obtained in the previous subsection tell us that we have designed a method that correctly provides fair recommendations. It is a simple, functional and easy to implement machine learning approach. Nevertheless, it has some drawbacks:

\begin{itemize}
    \item Choosing the adequate parameter alpha requires a fine-tuning process.
    \item Since the parameter alpha sign (less than or greater than zero) depends on the minority or non-minority nature of the recommended user, this recommendation method can only be applied to users with associated demographic information.
\end{itemize}

This subsection provides a \ac{DL} approach that works without the above drawbacks. This method only needs the parameter \(\beta\): it is used to select the accuracy vs. fairness balance. The \(\beta\) range is \(\left [0,1\right ]\), whether 0 means 100\% fairness and 0\% accuracy, and 1 means 100\% accuracy and 0\% fairness. As it can be seen, to choose a \(\beta\) value is straightforward and intuitive. Moreover: the chosen \(\beta\) value does not change when the user is a minority one or he is not.

\begin{figure}[ht]
    \centering
    \includegraphics[width=\textwidth]{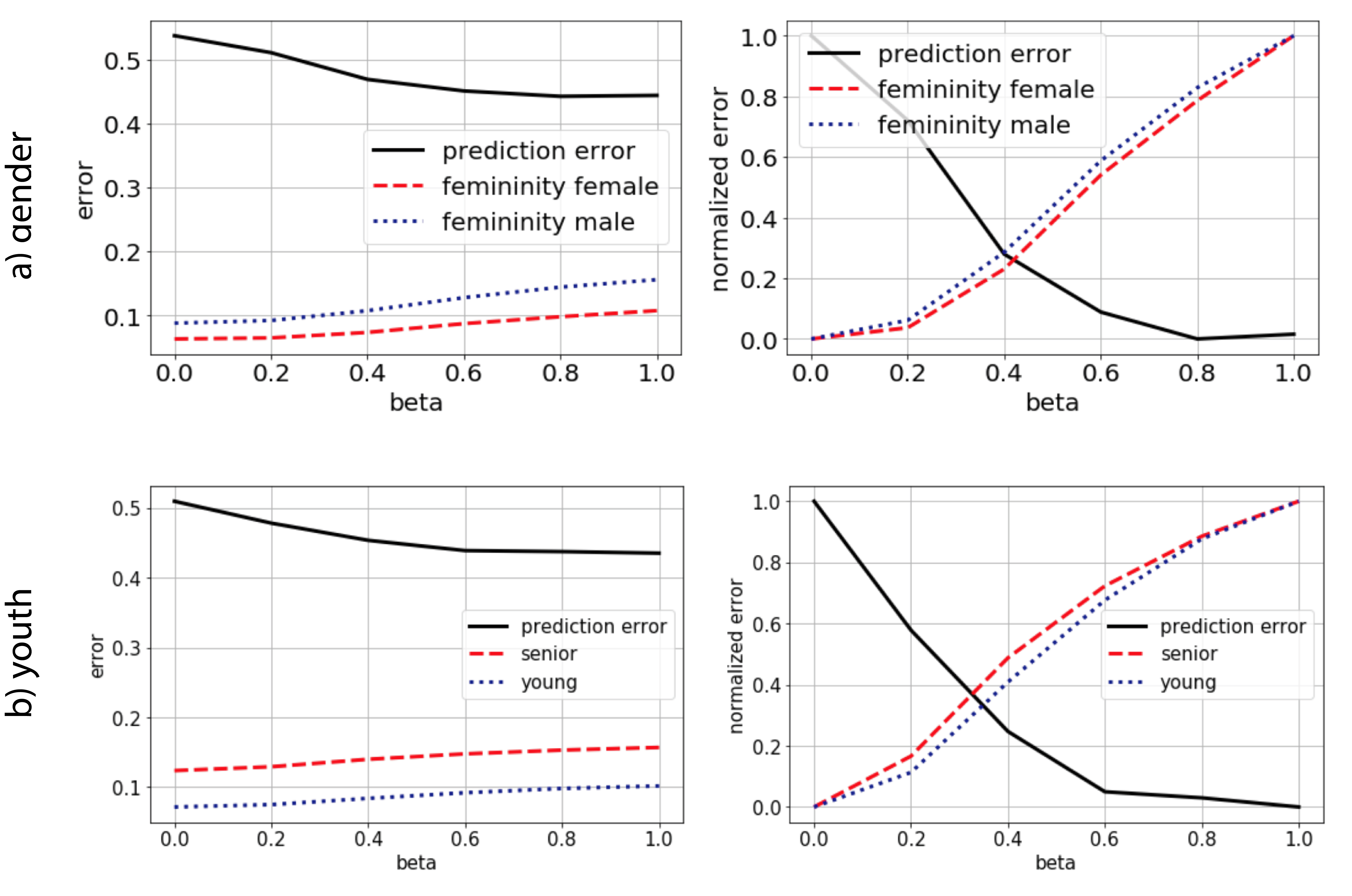}
    \caption{Recommendation results using the proposed \ac{DL} approach. y axis: averaged \(N=10\) recommendations error (normalized in the right graphs); x axis: \(\beta\) balance between fairness and accuracy (0.0 means 100\% fairness and 0\% accuracy, and 1.0 means 100\% accuracy and 0\% fairness).}\label{fig:rec-results}
\end{figure}

The proposed \ac{DL} recommendation method explained in \cref{sec:materials} returns the results shown in \cref{fig:rec-results}. Graphs on the left of the figure contain the main information. Graphs on the right are \(\left [0,1\right ]\) scaled to find the optimum accuracy vs. fairness balances. The averaged error of the recommendations (\cref{eq:26}) is plotted using black lines. Dotted and dashed lines show the minority errors (\cref{eq:27}); that is: the distance between the minority value of each recommended user (\textsc{UM}) and the average of the minority values (\textsc{IM}) of their N recommended items. We are looking for recommended items in the minority range of the user; e.g. if a user (male or female) has an \(UM=0.7\) (quite masculine), recommended items near \(IM=0.7\) are the fairest ones, and they generate a low minority (‘femininity’) error.

‘Gender’ results are shown in the top-left graph of \cref{fig:rec-results}: as expected, accuracy increases (error decreases) as \(\beta\) increases (more importance to accuracy). The price to pay for this accuracy improvement is the simultaneous increase in the fairness error values. As \(\beta\) decreases (more importance to fairness), the opposite happens: higher prediction errors and lower fairness errors. ‘Youth’ results are shown in the low-left graph of \cref{fig:rec-quality}: curve trends are similar to the ‘gender’ results. Graphs on the right of \cref{fig:rec-quality} show the same results by using a normalized y axis: in this way we can find the optimum \(\beta\) values to balance accuracy and fairness in the recommendation task. To optimize results in this experiment, it is necessary to choose a \(\beta =0.4\) value: a balanced selection, something scored to the fairness objective. This result tells us that the balanced option (\(\beta =0.5\)) can be the default one.

\section{Conclusions}\label{sec:conclusions}

Attending to the obtained results, it is understood that designing methods to improve \ac{CF} fairness is not a simple task, but it is possible to take it out. Due to the fact that an appreciable proportion of minority and non-minority users share preferences it is necessary to make use of modern machine learning approaches in order to  make fair recommendations not only to the ‘purest’ minority or non-minority users, but also to the users that mix some proportion of minority and non-minority preferences.

State of the art shows a lack of \ac{DL} approaches to tackle fairness in \ac{RS}, probably due to the neural networks black box model. The proposed method in this paper relies on an original loss function and input data to balance fairness and accuracy. This method combines several abstraction levels and it can serve as baseline to \ac{DL} future works in the field. An original architecture is provided, where machine learning and \ac{DL} models are combined to obtain balanced accuracy vs. fairness recommendations. The architecture is based on two basement levels: statistical and machine learning, that provide the necessary information to train the \ac{DL} model which constitutes the third architectural level. The proposed \ac{DL} method provides a modern approach to tackle fairness in \ac{RS}. We can easily balance accuracy and fairness, or we can automatically select the optimum trade-off. That is to say: the proposed method manages the inherent loss of accuracy when fairness is increased. Additionally, once the neural network is trained using demographic information, it can predict and recommend to users whose demographic information is unknown.  

Results show adequate trends in the tested quality measures: improvement in fairness at the cost of an expected worsening in accuracy. The proposed machine learning-based heuristic approach and the \ac{DL} model return similar quality results. Nevertheless, the proposed \ac{DL} method does not need demographic information in the recommendation feed-forward process. It also is able to better balance and automatically balance fairness and accuracy.

Proposed future works are: a) architecture simplification, by removing the \ac{MF} and transferring its functionality to the \ac{DL} model, b) items and users minority indexes redefinition to better catch the minority versus non-minority differences, c) testing the methods behaviour in a variety of CF datasets, d) extending the experiments to different demographic groups (nationality, profession, studies), and e) testing the architecture on not demographic groups (users that share minority preferences).

\bibliographystyle{abbrv}
\bibliography{fairness}

\end{document}